\documentclass[journal]{IEEEtai}

\usepackage[colorlinks,urlcolor=blue,linkcolor=blue,citecolor=blue]{hyperref}

\usepackage{color,array}
\usepackage{graphicx}
\usepackage{amsmath}
\usepackage{amsfonts}
\usepackage{booktabs} 
\usepackage{makecell} 
\usepackage{multirow} 
\usepackage{soul} 
\usepackage{hyperref}

\begin{document}

\title{Att\textsuperscript{2}CPC: Attention-Guided Lossy Attribute Compression of Point Clouds}

\author{
Kai~Liu,~\IEEEmembership{}
Kang You,~\IEEEmembership{}
Pan Gao,~\IEEEmembership{}
and~Manoranjan Paul~\IEEEmembership{}
\thanks{This work was supported by  the Natural Science Foundation of China under Grant 62272227. \emph{(Corresponding author: Pan Gao.)}}
\thanks{Kai~Liu, Kang You, and Pan Gao are with the College of Computer Science and Technology, Nanjing University of Aeronautics and Astronautics, Nanjing 211106, China, (E-mail: liu-kai@nuaa.edu.cn, youkang@nuaa.edu.cn, pan.gao@nuaa.edu.cn).}
\thanks{Manoranjan Paul is with School of Computing and Mathematics, Charles Sturt University, Bathurst, NSW 2678, Australia (email: mpaul@csu.edu.au).}
}

\markboth{Journal of IEEE Transactions on Artificial Intelligence, Vol. 00, No. 0, Month 2020}
{First A. Author \MakeLowercase{\textit{et al.}}: Bare Demo of IEEEtai.cls for IEEE Journals of IEEE Transactions on Artificial Intelligence}

\maketitle

\begin{abstract}
With the great progress of 3D sensing and acquisition technology, the volume of point cloud data has grown dramatically, which urges the development of efficient point cloud compression methods. In this paper, we focus on the task of learned lossy point cloud attribute compression (PCAC). We propose an efficient attention-based method for lossy compression of point cloud attributes leveraging on an autoencoder architecture. Specifically, at the encoding side, we conduct multiple downsampling to best exploit the local attribute patterns, in which effective \textcolor{black}{External Cross Attention (ECA)} is devised to hierarchically aggregate features by intergrating attributes and geometry contexts. At the decoding side, the attributes of the point cloud are progressively reconstructed based on the multi-scale representation and the zero-padding upsampling tactic. To the best of our knowledge, this is the first approach to introduce attention mechanism to point-based lossy PCAC task. We verify the compression efficiency of our model on various sequences, including human body frames, sparse objects, and large-scale point cloud scenes. Experiments show that our method achieves an average improvement of 1.15 dB and 2.13 dB in BD-PSNR of Y channel and YUV channel, respectively, when comparing with the state-of-the-art point-based method Deep-PCAC.  Codes of this paper are available at \href{https://github.com/I2-Multimedia-Lab/Att2CPC}{https://github.com/I2-Multimedia-Lab/Att2CPC}.
\end{abstract}

\begin{IEEEImpStatement}
Point cloud has been widely used in numerous fields, such as autonomous driving, Augmented Reality/Mixed Reality (AR/MR), robotics, etc, for the immersive and interactive applications. However, due to the inherent limitations of disk storage space and network bandwidth, the point cloud data need to be efficiently compressed. In this paper, we propose a novel Point Cloud Attribute Compression (PCAC) method that designs an attention-based network to exploit the correlation between geometry and attribute information, so as to further improve the performance of learning-based compression models. This paper also provides the community a novel perspective on tackling point cloud processing tasks, especially those involving joint analysis of geometry and attribute.
\end{IEEEImpStatement}


\section{Introduction}

\IEEEPARstart{P}{oint} cloud is a set of three-dimensional (3D) points with optional attributes such as color, normal, reflectance, etc. Due to its flexibility in representing 3D objects and scenes, point cloud has been widely used in many fields, such as autonomous driving \cite{geiger2012we}, Augmented Reality/Mixed Reality (AR/MR) \cite{pan2006virtual}, and digital city \cite{mekuria2016design}. With the significant advancements in 3D sensing and acquisition technologies, the size of point clouds has grown rapidly, posing significant challenges to the efficient transmission and storage of point cloud data.

\subsection{Research Background}

In the past several years, the Moving Picture Experts Group (MPEG) has introduced two Point Cloud Compression (PCC) standards: the Video-based Point Cloud Compression (V-PCC) and the Geometry-based Point Cloud Compression (G-PCC) ~\cite{graziosi2020overview,schwarz2018emerging}.
V-PCC first projects 3D point cloud sequences into a 2D plane, and then uses ready-made video codecs (e.g., HEVC~\cite{HEVC} or VVC~\cite{VVC}) to compress the projected video \cite{sullivan2012overview}. On the contrary, G-PCC encodes the point cloud frame directly in 3D space. It first adopts an octree structure to represent the point cloud geometry, and uses Predicting/Lifting Transforms or the Region-Adaptive Hierarchical Transform (RAHT~\cite{RAHT}) for attribute coding. 

With the successful application of Deep Neural Networks (DNNs) in 2D image/video compression techniques~\cite{balle2016end,ding2021advances,lu2023deep,TAI_compression}, both MPEG and Joint Photographic Experts Group (JPEG) have launched explorations for Artificial Intelligence (AI) based PCC solutions. However, despite the abundance of emerging learning-based PCC methods~\cite{zhang2022transformer,you2022ipdae,you2021patch,he2022density,wang2021multiscale,wang2022sparse,GRNet,deepPCAC,sparsePCAC,zhang2023yoga,ANF}, the majority of them are focused on the Point Cloud Geometry (PCG) compression~\cite{zhang2022transformer,you2022ipdae,you2021patch,he2022density,wang2021multiscale,wang2022sparse,GRNet}, while only a limited number of methods are available for attributes~\cite{deepPCAC,sparsePCAC,zhang2023yoga,ANF}. Nevertheless, the attribute information plays an essential role in practical applications (e.g., telecommunication, heritage archive, etc), which urges the development of learning-based Point Cloud Attribute Compression (PCAC) techniques.


\subsection{Current Challenges}

Compared with the relatively continuous and smooth appearance of point cloud geometric surfaces, point cloud attribute information exhibits stronger discontinuities and vagaries, posing significant challenges to lossy PCAC task.


The first challenge lies in how to effectively exploit geometry-to-attribute correlation. Previous methods, whether voxel-based~\cite{sparsePCAC,zhang2023yoga,ANF} or point-based~\cite{deepPCAC}, predominantly employed convolutions for the direct extraction of high-level attribute features, often overlooking the supplementary role of geometric information. Take the previous work Deep-PCAC~\cite{deepPCAC} as an example, it is the first one that adopts a point-based model and introduces an end-to-end variational autoencoder into lossy attribute compression. However, Deep-PCAC only considers the attribute features of points in convolution and aggregation, while neglecting the importance of the joint information between geometric and attribute patterns.

Another challenge emerges from the preprocessing step of voxelization, which can be interpreted as the process of quantizing the floating-point geometry coordinates to grid-based representation~\cite{graziosi2020overview}. However, this quantization process brings irreversible distortion. As point models advance in geometric compression~\cite{zhang2022transformer,you2022ipdae,huang20223qnet}, voxel-based PCAC techniques~\cite{wang2022sparse,zhang2023yoga,ANF} becomes unsuitable, since the process of voxelization may further deteriorates the geometry details reconstructed by point-based methods. Moreover, voxelization requires manually defined space precision, necessitating a thorough grasp of the spatial distribution of input point clouds. This makes current voxel-based models extremely challenging to automatically adapt to various point cloud scales and sparsity.

In recent years, the Transformer~\cite{vaswani2017attention} has shown significant effectiveness in numerous computer vision tasks, leading researchers to leverage it for point cloud analysis~\cite{guo2021pct,zhao2021point,zhang2022transformer,liu2022pcgformer,li2023proxyformer}. The core mechanism of Transformer is attention modeling, which has the potential to fuse the attribute features with geometry position embeddings, and has permutation invariance, that is, it is not affected by the order of points. Point-based transformer structures, such as Point Cloud Transformer (PCT)~\cite{guo2021pct} and Point Transformer (PT)~\cite{ptv2,PTv3}, also pioneer point-based attention mechanisms that  operate directly on the points. These characteristics make point-based attention mechanism very suitable for dealing with irregularly distributed point cloud data, motivating us to explore an effective attention-guided PCAC approach that meets the above-mentioned challenges.

\subsection{Our Approach}

In this paper, we propose an attention-guided point cloud attribute compression framework dubbed Att\textsuperscript{2}CPC. To the best of our knowledge, this study is the first exploration to integrate Transformer into lossy PCAC task. The proposed approach adopts a point-based autoencoder pipeline with geometry-based multiscale representation. Specifically, at the encoding side, we conduct multiple downsampling to best exploit the local attribute patterns, in which effective {External-Cross Attention (ECA)} is devised to hierarchically aggregate features by intergrating attributes and geometry contexts. At the decoding side, the attributes of the point cloud are progressively reconstructed based on the multi-scale representation and the zero-padding upsampling tactic. Experiments show that our method attains superior compression performance when comparing with previous learning-based methods. 


Main contributions of this paper are as follows:

\begin{itemize}

    \item This paper presents the first Transformer-based point cloud attribute compression framework. By arming the multi-scale representation with the point-based Cross Attention mechanism in an autoencoder fashion, our encoder yields a compact representation of point cloud attribute features.

    \item An effective \textcolor{black}{External Cross Attention (ECA)} mechanism is devised to effectively exploit the correlations between point cloud geometries and attributes. A novel zero-padding upsampling tactic is developed to facilitate the adaptive learning of the attributes of the upsampled points.
    
    \item Our method demonstrates superior compression efficiency when comparing with other learning-based methods. Since our method directly operates on points, it can streamline the preprocessing and avoid the distortions introduced by voxelization.
    
\end{itemize}


\section{Related Work}
\subsection{Point Cloud Analysis}

\subsubsection{Voxel-based method}
Since the points do not have a specific order or a grid structure, voxelization is a commonly used preprocessing step. That is, the point cloud is quantized into cubes first, and then 3D convolution is employed to extract a latent feature representation \cite{maturana2015voxnet,riegler2017octnet}. However, voxelization lacks the flexibility to express the intricate topological structure of points, owing to the loss of detailed information caused by irreversible distortion. 


\subsubsection{Point-based method}
Thanks to PointNet \cite{qi2017pointnet} and PointNet++ \cite{qi2017pointnet++}, point-based methods that directly operate on points have been widely studied and examined \cite{guo2021pct,guo2022attention,hui2021pyramid,TAI_PC_Registration}. Compared with voxel-based methods, point-based pipeline can better preserve the original topological structure of point clouds, due to the avoidance of voxelization process. With the major breakthrough of Transformer \cite{vaswani2017attention} in the field of natural language processing and 2D computer vision, researchers have introduced Transformer into point cloud tasks and have achieved exciting results \cite{guo2021pct,guo2022attention,hui2021pyramid}. To the best of our knowledge, there is no research that incorporates attention into point model-based PCAC task, which presents an opportunity for further exploration.
{
Nevertheless, compared to other point cloud analysis tasks, Transformer-based PCAC faces several major challenges when encoding:
1) The huge number of points of input point clouds makes the Transformer perform globally computationally infeasible and difficult to run on general-purpose computing machine like ours.
2) It is very difficult to extract color distribution features based on disordered point cloud geometric information.
In response to the above problems, we propose to gradually extract local features of point clouds in stages and design a novel \textcolor{black}{ECA} module to efficiently extract color features.
}

\subsection{Point Cloud Attribute Compression}
Attribute compression involves reducing redundancy between point attributes, which can be divided into conventional transform-based methods and learning-based methods.

\subsubsection{Transform-based methods}
Following the general idea of image compression, many works focus on designing geometrically dependent transformations to convert attribute signals into specific frequency domains followed by quantization and entropy coding. For instance, due to the strong sparse representation ability of Graph Fourier Transform (GFT), some methods \cite{shao2018hybrid,zhang2014point,song2022rate} try to construct graphs with the geometric coordinates of points and apply GFT to transform attribute signals. However, the frequent eigenvalue decomposition of the Laplacian matrix results in high computational cost.
To alleviate this problem, region adaptive hierarchical transform (RAHT) is proposed in \cite{RAHT}, which utilizes a hierarchical sub-band transform that resembles an adaptive variation of the Haar wavelet. RAHT can significantly reduce computational complexity and achieve faster processing speed through the adaptive hierarchical transform strategy, and it has adopted in G-PCC on account of its superior efficiency. In addition, The predicting and lifting transforms are also the core algorithms of G-PCC for point cloud attribute compression.

\subsubsection{Learning-based methods}
Learning-based point cloud compression has received increasing attention in recent years. 3DAC \cite{fang20223dac} proposes a deep entropy model that can estimate the probabilities of the transformation coefficients of point cloud attributes. IPPCAC \cite{9949997} employs region-adaptive layered transformation and adaptive neighborhood selection modules for compressing attribute information within 3D point clouds. CNeT \cite{kaup2023lossless} and MNeT \cite{10095385} utilize learned conditional probability models to achieve lossless point cloud geometry and attribute compression. \cite{Wang2023LosslessPC} utilizes interconnections among different scales, groups, and color attributes within point clouds to achieve effective lossless compression. Similarly, \cite{Wei2022ContentAdaptiveLO} utilizes point cloud level-of-detail hierarchies and prediction error models to achieve content-adaptive lossless compression of point cloud attributes. 
Deep-PCAC \cite{deepPCAC} employ lossy compression techniques to compress the attribute information of 3D point clouds, which is a point-based model. 
However, Deep-PCAC neglects the dependence of attribute compression on geometry context, which, nevertheless, is very important for attribute compression performance. Further, it simply employs MLP and pooling to aggregate the feature with neighboring points, which cannot fully exploit the inherent correlation between neighboring points. In this work, we also consider the point-based compression paradigm. But contrary to Deep-PCAC, we take one step further.  We incorporate attention mechanism into attribute compression, which considers both cross attention between geometry context and attribute, and self-attention among neighboring points.


\section{Methodology}

\subsection{Overview}
\begin{figure*}[t]
\centering
\includegraphics[width=1.0\textwidth]{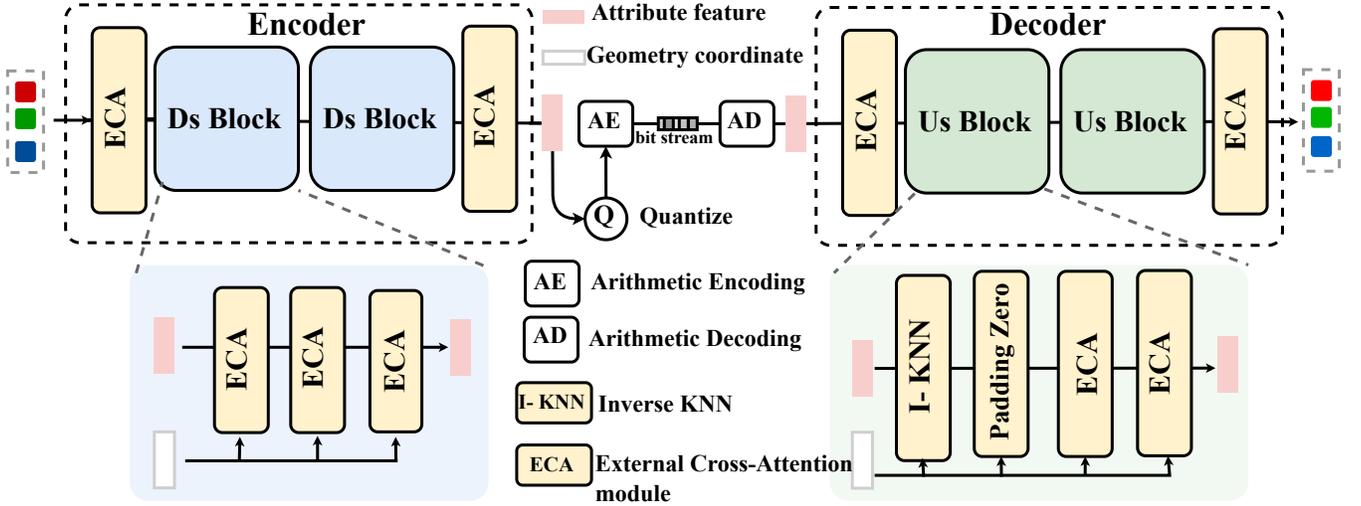}
\caption{The overall framework of our approach. \textcolor{black}{ECA} represents the \textcolor{black}{External Cross Attention} Module that abstracts color features based on geometry context. FPS denotes the Farthest Point Sampling that extracts uniform sampling points from the original point cloud. AE and AD refers to arithmetic encoding and arithmetic decoding, respectively.}
\label{fig:overview}
\end{figure*}

We present a lossy point cloud attribute compression model in the context of  geometry losslessly transmitted. Since every attribute set needs to reside on a point geometry, the original geometry information is utilized to assist the network to produce compact representation of attribute patterns. As shown in Fig. \ref{fig:overview}, our proposed method utilizes an auto-encoder pipeline in a multi-scale fashion, which comprises a downsampling-based encoder, an entropy engine, and an upsampling-based decoder. During the encoding, the \textcolor{black}{External Cross Attention (ECA)} module is deployed in stacked Down-sampling Blocks to generate compact attribute features. These compact latent features are then quantized and encoded by an entropy engine, resulting in a compressed bit stream. The decoder, serving as the inverse of the encoder, using stacked upsampling blocks to reconstruct the attributes. Specific design of the network will be elaborated on in the following subsections.

\subsection{Encoder}
Our encoder extracts attribute features by aggregating and downsampling the point cloud multiple times. The \textcolor{black}{External Cross Attention (ECA)} module is used to efficiently extract attributes from neighbor points, which allows our network to learn color patterns in detail based on joint geometric information. The Farthest Point Sampling (FPS) is used in our encoder to obtain more compact representations of the local details.

\subsubsection{\textcolor{black}{External Cross Attention (ECA)}} 
We assume that there exists a strong correlation between attributes and coordinates, since points of a point cloud surface are continuously distributed in Euclidean space, two points closer to each other tend to have similar attribute values. Owing to the cross attention\cite{feng2021accelerated} that effectively extract the correlation between multi-modal information, a novel cross attention-based feature fusion module is devised, aiming at fully mining the correlation between attributes and coordinates.

\begin{figure}[t]
\centering
    \includegraphics[width=1.0\linewidth]{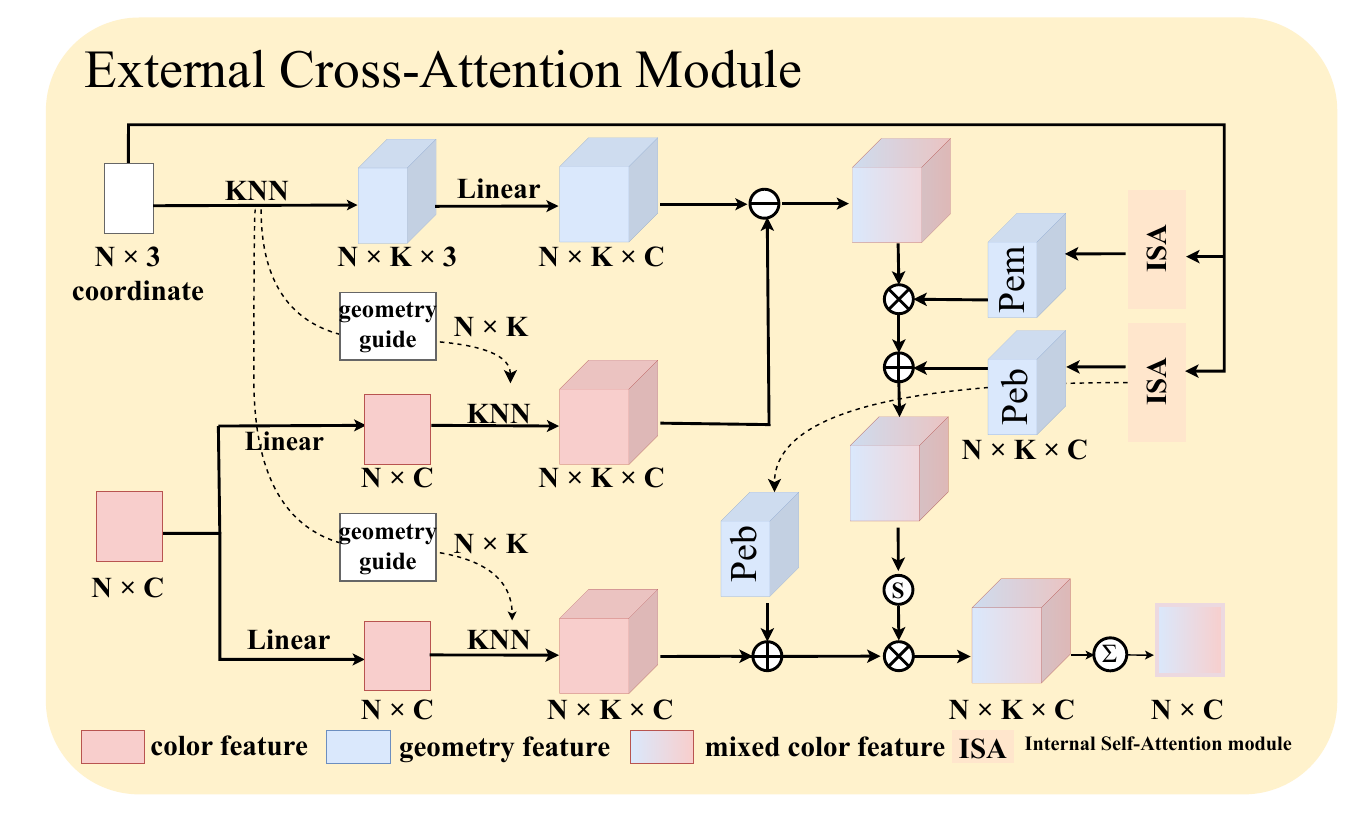}
\caption{Structure of \textcolor{black}{External Cross Attention module (ECA)}.}
\label{fig:PCA}
\end{figure}

\begin{figure}[h!]
\centering
\includegraphics[width=1.0\linewidth]{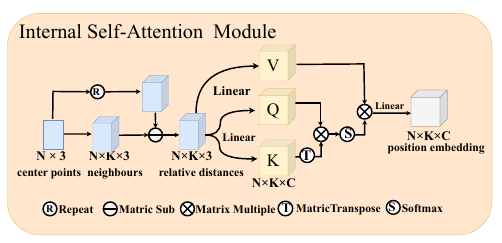}
\caption{Structure of Internal Self-Attention module (ISA).} 
\label{fig:SSA}
\end{figure}

As shown in Fig. \ref{fig:PCA}, in our \textcolor{black}{ECA} module, two \textcolor{black}{Internal Self-Attention ({ISA})} modules are utilized to facilitate local position embeddings. This nested dual-attention mechanism allows us to fully exploiting the correlation of geometry and attribute signals. Specifically, we input the point cloud coordinates tensor with shape $N \times 3$ and regard each point $p_{i}$ in the point cloud $\mathcal{P}$ as the center point, then search for the $k$ neighbor points $knn\left ( p_{i}  \right )$ closest to the center point $p_{i}$ in the point cloud, resulting in the local sensing area tensor of each point with shape $N \times K \times 3$.

Then, we calculate the relative position $\Delta p_{ij}$ of these neighbor points relative to the center point, i.e., 
\begin{equation}
    knn\left ( p_{i}  \right ) = {\left \{p_{i1},p_{i2},...,p_{ik}\right \}}, \quad p_{i} \in \mathcal{P}
\end{equation}
\begin{equation}
    \Delta p_{ij} = p_{ij}-p_{i}, \quad p_{ij}\in  knn\left ( p_{i}  \right )
\end{equation}
where $\mathcal{P}$ refers to the input point set of each input down-sampling layer, $k$ refers to the number of neighbor points.

Then, we send these relative position coordinates $(N \times K \times 3)$ to the \textcolor{black}{ISA} module to generate position embeddings, including position embedding multiplier ($Pem$) and position embedding bias ($Peb$) with the same shape $N \times K \times C$. Mathematically, denote the relative position $\Delta p_{ij}$ within a local neighbor area as $X$, the process of generating position embeddings (\textcolor{black}{ISA}) can be represented as:
\begin{equation}
    {\textcolor{black}{ISA}}\left(X\right) = \text{Softmax}\left (\frac{{\phi}_{Q}(X){\phi}_{K}(X)^{T}}{\sqrt{d}}\right){\phi}_{V}(X)
\end{equation}
\begin{equation}
    Pem = {\phi}_{Pem}({\textcolor{black}{ISA}}_{Pem}(X))
\end{equation}
\begin{equation}
    Peb = {\phi}_{Peb}({\textcolor{black}{ISA}})
\end{equation}
where ${\phi}$ represents the multi-layer perceptron (MLP), $d$ represent the dimension of the mapped features from MLPs. 

Next, the obtained positional encodings $Pem$ and $Peb$ are fused into the color information aided by our devised \textcolor{black}{ECA} module, as detailed below.

In \textcolor{black}{ECA} module, we input the geometric coordinates (with shape $N \times 3$) of the point cloud and extract the K nearest neighbors of each point, obtaining point-wise local sensing area ($N \times K \times 3$) and geometric indices ($N \times K$) that guide the sampling of color features. Then, we use a linear layer to expand the dimension of the local sensing area with shape from $N \times K \times 3$ to $N \times K \times C$. For the input color features ($N \times C$), we first use two parallel linear layers, and then sample two color feature neighborhoods from the local sensing area according to the geometric indices. Inspired by \cite{ptv2}, we use the geometric features (Q) in the local sensing area to subtract the color features (K), trying to obtain the mapping relationship between geometry and color, and fuse the position embeddings (PEM and PEB) by multiplication and addition, resulting in a score matrix with shape $N \times K \times C $. Another color feature neighborhood tensor is added with PEB, and then multiplied with the score matrix, resulting in a mixed color feature, and finally the neighborhood features are aggregated by the sum operation.
This process can be explicitly expressed as follows.
\begin{equation}
    \mathcal{S} =\text{Softmax}\left(\left(\psi_{K}(\mathcal{F}_{in})-\psi_{Q}(X)\right)\odot Pem +Peb\right)
\end{equation}
\begin{equation}
{\textcolor{black}{ECA}}\left(\mathcal{F}_{in},Pem,Peb\right)=\text{Sum}\left(\left(\psi_{V}(\mathcal{F}_{in})+Peb\right)\odot \mathcal{S}\right)
\end{equation}
where $\mathcal{F}_{in}$ represents the input attribute features, ${\psi}$ represents the multi-layer perceptron (MLP), $\mathcal{S}$ represents the scores output by an \text{Softmax} function, \text{Sum} represents the point-wise sum operation, and $\odot$ denotes the Hadamard product. In this instance, $\mathcal{F}_{in} \in \mathbb{R}^{N \times K \times C}$.


\subsubsection{Downsampling Block}
In the down-sampling block, we first stack the \textcolor{black}{ECA} module to obtain the enhanced attribute features. The farthest point sampling (FPS) \cite{qi2017pointnet++} is used to uniformly downsample the input point set from the original scale to generate compact point cloud representations. FPS first randomly picks a point in the point cloud and iteratively searches for the farthest point until a specified number of points is obtained. It can produce uniform skeleton points of the original point cloud, which helps us retain the overall contour information of the point cloud during downsampling.

\subsection{Quantization and Bitrate Estimation}
An uniform scalar quantizer is used in this work following the widely-used soft quantization strategy \cite{balle2016end}. To be specific, a uniform noise ranging from -0.5 to 0.5 is added on the bottleneck features during training. Mathematically,
\begin{equation}
Q(f) = f + f_{noise} = f + \mathcal{U}(-0.5, 0.5) \\
\end{equation}
where $f$ refers to the feature output of the encoder, $Q$ represents the quantization operation, and $\mathcal{U}$ represents the noise generator which generates random noises from a uniform distribution of [-0.5, 0.5]. A rounding operation is applied to get the quantized feature values in test conditions. 

The expected code length (i.e., the bit rate $R$) is estimated by a cross entropy:
\begin{equation}
    R=\mathbb{E}_{f \in p_{f} }[-\log_{2}{ p_{\hat{f}}(Q(f)) }]
\label{eq:bitrate}
\end{equation}
where $p_{f}$ denotes the actual distribution of the features, and $p_{\hat{f}}$ is the entropy model.

\subsection{Decoder}
The task of the decoder is to reconstruct point cloud attributes from the low-dimensional attribute features. In this paper, an up-sampling scheme is used to bounce attribute patterns back, by using inverse K-nearest neighbors (KNN) algorithm, novel padding zero strategy, and \textcolor{black}{ECA} modules that restore attribute values.

\begin{figure}[t]
\centering
\includegraphics[width=1.0\linewidth]{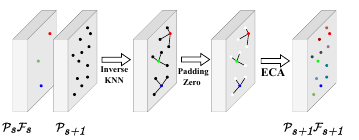}
\caption{Upsampling Block. The black points indicates that the attribute value of the points is to be recontrusted, and the white points indicates that the attribute value of the points is zero. $\mathcal{P}_s$ and $\mathcal{F}_s$ denotes the input point geometry coordinates and input features, respectively. $\mathcal{P}_{s+1}$ and $\mathcal{F}_{s+1}$ denotes the upsampled coordinates and features, respectively.}
\label{fig:Decoder}
\end{figure}

Our upsampling block takes the down-sampled sparse skeleton as the input and upsamples to a dense point set, as shown in Fig. \ref{fig:Decoder}. Mathematically, let $\mathcal{P}_{s}$ be the input coordinate set, and $\mathcal{F}_{s}$ be the corresponding features, then the output of our upsampling block $\mathcal{F}_{s+1}$ can be calculated as follows:
\begin{equation}
    \mathcal{F}_{s+1} = \text{UsBlock}\left( \mathcal{P}_{s},\mathcal{F}_{s},\mathcal{P}_{s+1} \right)
\end{equation}

In this paper, we devise a novel zero-padding strategy to interpolate feature values for the upsampled point anchor. Different from the up-sampling scheme in Deep-PCAC \cite{9447226} and MS-GAT \cite{9767661} that use the distance-weighted interpolation of neighboring points, which may lead to a constrained network prediction of attribute values, we directly initializing the attribute feature to zero to provide a greater degree of freedom for the neural network. 

Specifically, for each point in the dense point cloud anchor $\mathcal{P}_{s+1}$, an inverse KNN is first utilized to find its nearest neighbor in the input point set $\mathcal{P}_{s}$. Let $x^{i}_{s+1}$ be the point in the dense anchor, $x^{j}_{s}$ be the nearest neighbor in the input point set, and $f^{j}_{s}$ be the feature of the point $x^{j}_{s}$, then our zero-padding strategy to generate $f^{i}_{s+1}$ can be described as follows:
\begin{equation}
    f^{i}_{s+1} = \begin{cases}
                      f^{j}_{s}, & \text{ if } \left \| x^{i}_{s+1} - x^{j}_{s} \right \|^2_2 = 0\\
                      \varrho,         & \text{ otherwise }
                    \end{cases}
\label{eq:zero_padding}
\end{equation}
where $\varrho$ represents the zero vector which has the same tensor shape with the feature $f^{j}_{s}$. From the condition of Eq. \ref{eq:zero_padding}, it can be seen that we only retain the features of the sparse input point set, while all the features of the added points will be filled as zero. After we attach feature values to the dense anchors, stacked \textcolor{black}{ECA} is utilized to further enhance the interpolated features.

Using zero-padding instead of distance-weighted interpolation has two merits. First, it runs faster than the conventional multi-neighbor interpolation since we only search for one nearest neighbor. Second, as opposed to rule-based interpolating, the network will actively control how it interpolates, which leads to a more adaptive solution.

\subsection{Loss function}
We train the model using a rate-distortion loss. In our approach, we define distortion as the attribute difference between the reconstructed point cloud and the original point cloud, the loss function is defined as follows:
\begin{equation}
    L(\hat{a}_{i},a_{i})=\sum_{i}^{} \left \| \hat{a}_{i} - a_{i}    \right \|_{2}^{2} + \lambda R   
\end{equation}
where $\hat{a}_{i}$ denotes the attribute values in the predicted point cloud, and $a_{i}$ denotes the attribute values in the original point cloud, and $R$ denotes the 
estimated bit rate (as shown in Eq. \ref{eq:bitrate}). $\lambda$ is used to balance the distortion and bit rate, which is set to different values to fit various bit-rate budgets.


 \begin{figure}[t]
\includegraphics[width=1.0\linewidth]{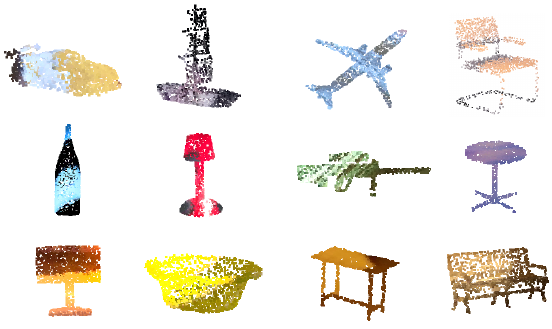}
\caption{Examples of our training set. Point clouds are synthesized from ShapeNet \cite{shapenet} and PCCD \cite{pccd}.}
\label{fig:training_set}
\end{figure}

\begin{figure}[t]
\includegraphics[width=1.0\linewidth]{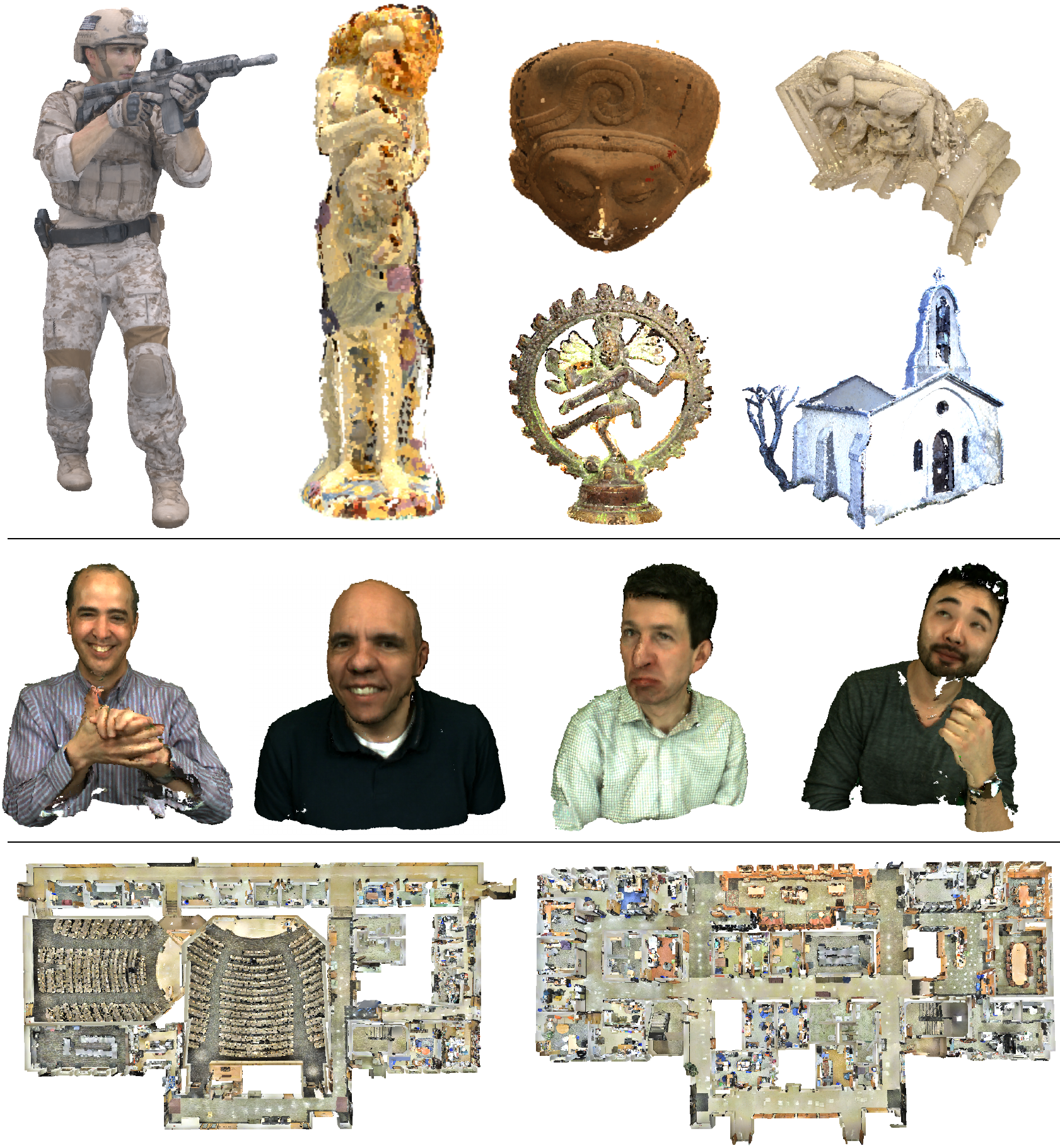}
\caption{Test sequences.}
\label{fig:testing_set}
\end{figure}

\section{Experiment}
\subsection{Datasets}
\textbf{Training.} Inspired by SparsePCAC \cite{wang2022sparse}, we utilize the following two datasets to synthesize a training set: the image aesthetics dataset PCCD \cite{pccd} and the point cloud geometry dataset ShapeNet \cite{shapenet}. We randomly map images from the PCCD dataset onto ShapeNet-generated point cloud geometry samples, which involves ShapeNet providing the geometric coordinate of each point and PCCD providing the attached attribute values. Examples of our synthetic point cloud can be found in Fig. \ref{fig:training_set}.

\textbf{Test.} To allow for fair comparison with the SOTA method (e.g., Deep-PCAC \cite{deepPCAC}), similar sequences are used for the tests in this paper, which includes ``Mask'', ``Frog'', ``Staue Klimt'', ``Shiva'', ``Soldier'', and ``House'' in the MPEG PCC Category 2 \cite{schwarz2018common}, ``Area2'' and ``Area4'' in the Stanford 3D Large-Scale Indoor Spaces (S3DIS) dataset \cite{armeni_cvpr16}, ``phil9'', ``sarah9'', ``andrew9'', and ``david9'' from Microsoft Voxelized Upper Bodies (MVUB) \cite{mvub}. The visualization of the test sequences is shown in Fig. \ref{fig:testing_set}, in which it can be observed that our test samples covered a wide range of application scenarios including dense human-body frames, sparse objects, and large-scale point cloud scenes.

Given the large number of points in the test samples (as shown in Tab. \ref{tab:point number}), which is impossible to take the entire point cloud as the input of our network, we divide each point cloud into smaller patches and compress them sequentially. In our experiment, each patch contains 2048 points, which aligns with the number of points of point clouds in our training set.

\begin{table}[t]
\centering
\caption{NUMBER OF POINTS OF THE TEST SEQUENCES.}
\begin{tabular}{c|c}  
\toprule 
Point cloud & Number of Points \\
\hline 
\verb|Egyptian_mask_vox12| & \raisebox{-0.3ex}{272,684} \\\hline
\verb|Frog_00067_vox12| & \raisebox{-0.3ex}{3,614,251} \\\hline
\verb|Staue_Klimt_vox12| & \raisebox{-0.3ex}{499,660} \\\hline
\verb|Shiva_00035_vox12| & \raisebox{-0.3ex}{1,009,132} \\\hline
\verb|soldier_vox10_0690| & \raisebox{-0.3ex}{1,089,091} \\\hline
\verb|House_without_roof_00057_vox12| & \raisebox{-0.3ex}{4,848,745} \\\hline
\verb|S3DIS_Area2| & \raisebox{-0.3ex}{47,315,372}  \\\hline
\verb|S3DIS_Area4| & \raisebox{-0.3ex}{43,470,014} \\\hline
\verb|andrew9_frame0000| & \raisebox{-0.3ex}{279,644}  \\\hline 
\verb|david9_frame0000| & \raisebox{-0.3ex}{330,797}  \\\hline 
\verb|phil9_frame0000| & \raisebox{-0.3ex}{370,798}  \\\hline 
\verb|ricardo9_frame0000| & \raisebox{-0.3ex}{214,656}  \\\hline 
\end{tabular}
\label{tab:point number}
\end{table}

\begin{figure*}[t]
\centering
\includegraphics[scale=0.38]{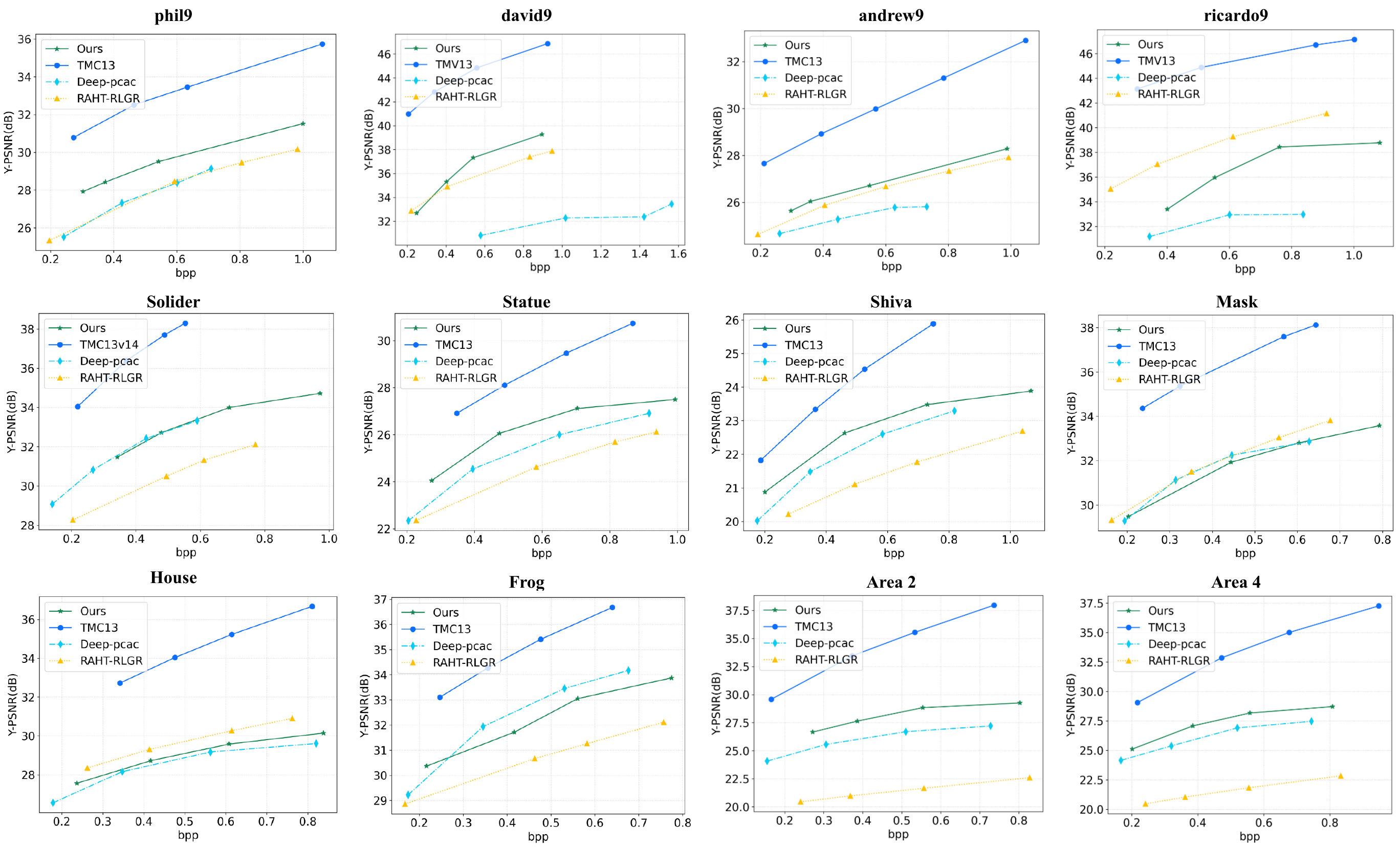}
\caption{Rate-distortion curves for our proposed method, TMC13, Deep-PCAC and RAHT-RLGR. Distortion is measured on the peak signal-to-noise ratio (PSNR) of the Y component.}
\label{fig:RD-curve}
\end{figure*}

\subsection{Implementation Details}
The YUV color space is used in this paper due to its statistical efficiency for lossy attribute compression. The sampling rate is set to 4 for each down-sampling and up-sampling layer. The feature dimensions of all hidden layers are set to 256. 

We perform our experiments using Python 3.9 and Pytorch 1.11. The Adam \cite{Adam} optimizer is used with a learning rate of 0.0005. The parameters $\beta_{1}$ and $\beta_{2}$ are set to 0.9 and 0.999, respectively. We train our model on one NVIDIA RTX2080Ti GPU with 120,000 steps for each bit-rate point. The $\lambda$ is set to $3\times10^{-4}$, $6\times10^{-4}$, $1\times10^{-4}$, and $8\times10^{-5}$ under different bit rates, with each model takes about 12 hours for training.

\begin{figure*}[t]
\centering
\includegraphics[scale=0.38]{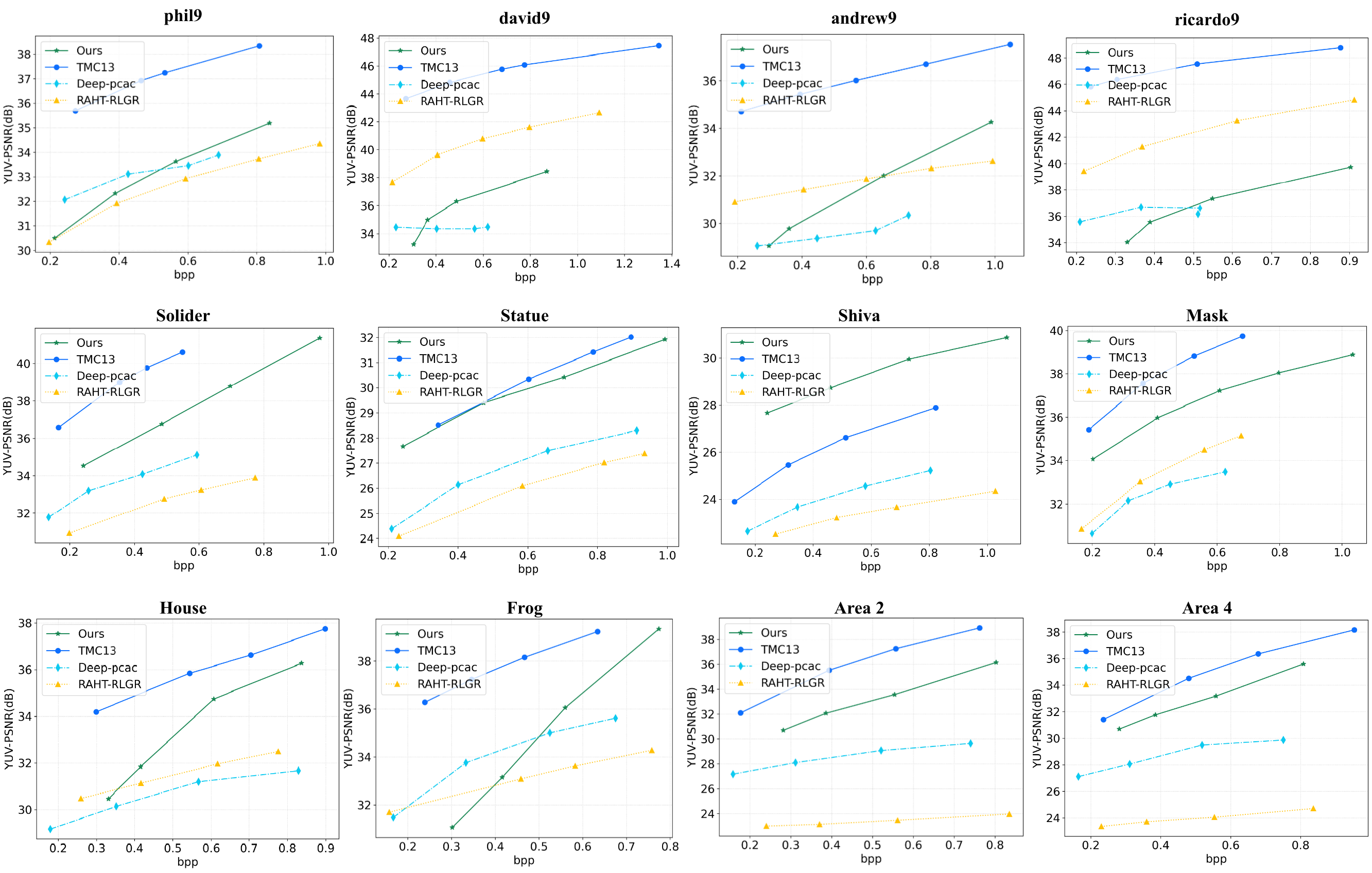}
\caption{We compare our method with Deep-PCAC, G-PCC, and RAHT-RLGR by plotting rate-distortion curves, and the measure of quality is based on the peak signal-to-noise ratio (PSNR) of the YUV component.}
\label{fig:RD-curve2}
\end{figure*}




\subsection{Objective Evaluation}
We focus on comparing the peak signal-to-noise ratio of the Y channel (Y-PSNR) since the luma channel accounts for the majority of compressed bit stream \cite{tian2017evaluation}. The Bits per point (Bpp) serves as the metric to measure the bit cost of compressed attribute values. 

Figure \ref{fig:RD-curve} shows the rate-distortion curves of our method and three other representative methods including G-PCC (TMC13) \cite{WG7PCC}, Deep-PCAC \cite{deepPCAC} and RAHT-RLGR \cite{RAHT}. Note that the SparsePCAC \cite{wang2022sparse} is not included for its unreleased source code. It can be seen from Fig. \ref{fig:RD-curve} that our method can match or surpass the compression performance of the Deep-PCAC and RAHT-RLGR on most point cloud samples, which demonstrates the superiority of our method. However, RAHT-RLGR usually performs better on point clouds with simple color distributions, which can be more efficiently exploited through Haar-wavelet transform. It can be observed that from ``ricardo9'', ''mask'', and ``house'', where RAHT-RLGR provides decent compression performance than ours. An effective model based on color decomposition and aggregation is one of the interesting topic in our future work.

We use BD-BR and BD-PSNR as metrics to compare the Y-PSNR of different methods, as shown in Tab. \ref{tab:result}. Our method provides 31.36\% BD-BR and 1.58dB BD-PSNR gains compared to rules-based RAHT-RLGR on average. When comparing the learning-based Deep-PCAC, our method achieves an average gains of 31.69\% and 1.43dB in terms of BD-BR and BD-PSNR, respectively. However, it should be noted that it is still difficult for point model-based solutions to meet up with G-PCC at present, which could be the inefficient probability distribution modeling during entropy coding. Hyper-prior \cite{balle2018variational} and auto-regressive techniques \cite{minnen2018joint,mentzer2018conditional} will be included in our future research. 

To present more comprehensive experimental results, the YUV-PSNR metric is included when comparing with Deep-PCAC. As shown in Tab. \ref{tab:YUV-PSNR-table}, our method provides significant improvements in objective visual quality, with an average of 148.13\% bit-rate reduction in BD-BR and 2.13dB gain in BD-PSNR, demonstrating the advanced performance of our method at chroma channels.

\setlength{\tabcolsep}{0.27em} 
\begin{table}[t]
\centering
\caption{QUALITATIVE COMPARISON BETWEEN G-PCC(TMC13), Deep-PCAC, and RAHT-RGLR. OUR PROPOSED METHOD SERVES AS THE ANCHOR. ABNORMAL BD-BR VALUES (E.G., \textgreater999\%) ARE DISCARDED FOR REASONABLE AVERAGED DATA, AND THEY ARE CROSSED FOR INDICATION.}
\label{table}
\begin{tabular}{c|c|c|c|c|c|c}
\toprule
\multirow{3}{*}{PC} & \multicolumn{2}{c|}{TMC13} & \multicolumn{2}{c|}{Deep-PCAC} & \multicolumn{2}{c}{ RAHT-RGLR} \\ 
\cline{2-7} 
 & \footnotesize{\makecell{BD-BR\\(\%)}} & \footnotesize{\makecell{BD-PSNR\\(dB)}} & \footnotesize{\makecell{BD-BR\\(\%)}} & \footnotesize{\makecell{BD-PSNR\\(dB)}} & \footnotesize{\makecell{BD-BR\\(\%)}} & \footnotesize{\makecell{BD-PSNR\\(dB)}} \\ 
\hline
Soldier  &+245.48 &-4.79   &-51.42 &+2.15  &-84.54   & +2.31\\ \hline
Mask     &+346.14 & -4.59  &+4.61  &-0.18  &+11.51   & +0.37\\ \hline
Frog     &+132.48 &-3.13   &-20.05 &-0.70  &-66.31   &+1.28\\ \hline
House    &\st{+1456} & -5.32   &-9.71  &+0.24  & +24.47   & -0.65\\ \hline
Shiva    &+80.39   & -1.41 &-18.80 &+0.52  &-108.60  & +1.54\\ \hline
Staue    &+100.79  &-2.26  &-20.41 &+0.86  &-66.88   & +1.84\\ \hline
Phil9    &+194.16  &-3.49  &-26.82 &+0.88  &-34.86   & +1.32\\ \hline

Ricardo9 &\st{+877081} &-8.79   &\st{+2962} &+3.34   &+37.38   &-2.87\\ \hline

Andrew9  &+288.22 & -3.24 &-39.68 &+1.14   &-10.28   & +0.28\\ \hline
David9   &+712.89 &-7.89  &-92.79 &+6.09   &-15.48   & +0.85\\ \hline
Area2    &+522.77 &-6.52  &-47.19 &+1.72   &\st{-475701}  & +6.78\\ \hline
Area4    &+354.02 & -5.1  &-26.31 &+1.06   &\st{-2450}    & +5.94\\ \hline
\
\textbf{Average} & +403.11 &-4.31 & -38.60 & +1.15 & -22.65 & +1.11\\ 
\bottomrule
\end{tabular}
\label{tab:result}
\end{table}

\begin{table}[t]
\centering
\caption{BD-PSNR(dB) and BD-BR(\%) OF YUV COMPOSITE COMPONENTS COMPARED TO Deep-PCAC. SIMILARLY, THE ABNORMAL  VALUES ARE CROSSED.}
\begin{tabular}{c|c|c}  
\toprule 
Point Cloud& BD-BR(\%)  &BD-PSNR(dB)\\ \hline
Soldier &-72.56  & +1.81  \\ \hline
Mask    & -328.91 &+3.27   \\ \hline
Frog    & +26.03 & +0.49  \\ \hline
House   & -46.92 & +2.57  \\ \hline
Shiva   & \st{-2247.90} & +4.62  \\ \hline
Statue  & -196.81 &+2.89   \\ \hline
Andrew9 & -57.82 & +0.38  \\ \hline
Ricardo9 & -9.09   & +0.35 \\ \hline
David9 & -66.76  & +1.27  \\ \hline
Phil9  & -26.49  &  +0.38  \\ \hline
Area2  & - 349.56  & +4.02  \\ \hline
Area4  & -500.53  & +3.45 \\ \hline
\textbf{Average}& -148.13 & +2.13     \\\bottomrule
\end{tabular}
\label{tab:YUV-PSNR-table}
\end{table}

\begin{figure*}[t!]
\centering
\includegraphics[width=1.0\linewidth]{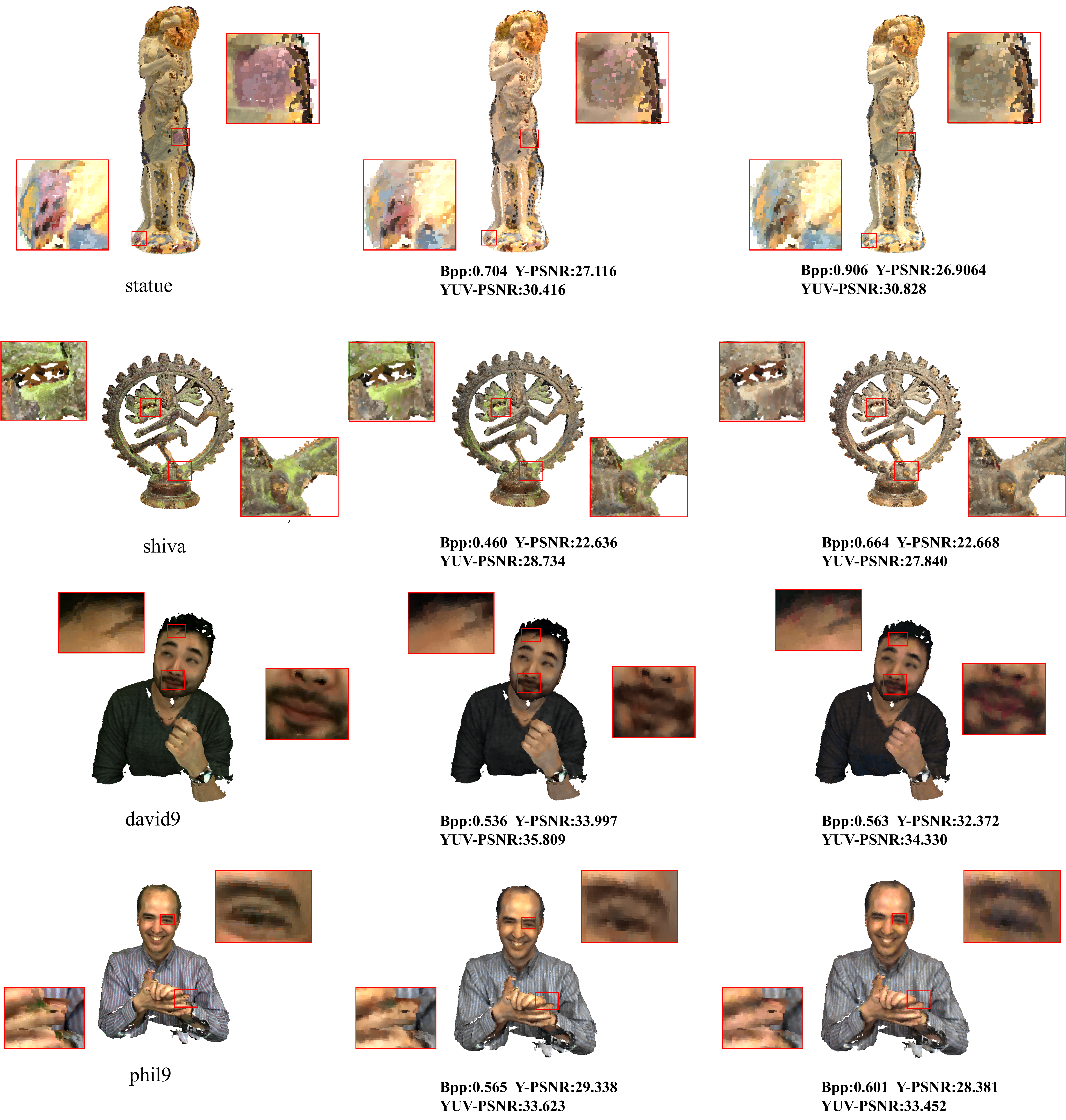}
\caption{Visual results of our method and Deep-PCAC. The left column is the original point cloud, the middle column is the point cloud reconstructed by our method, and the right column is the point cloud reconstructed by Deep-PCAC.}
\label{fig:visual}  
\end{figure*}

\begin{figure*}[h]
\centering
\includegraphics[width=1.0\linewidth]{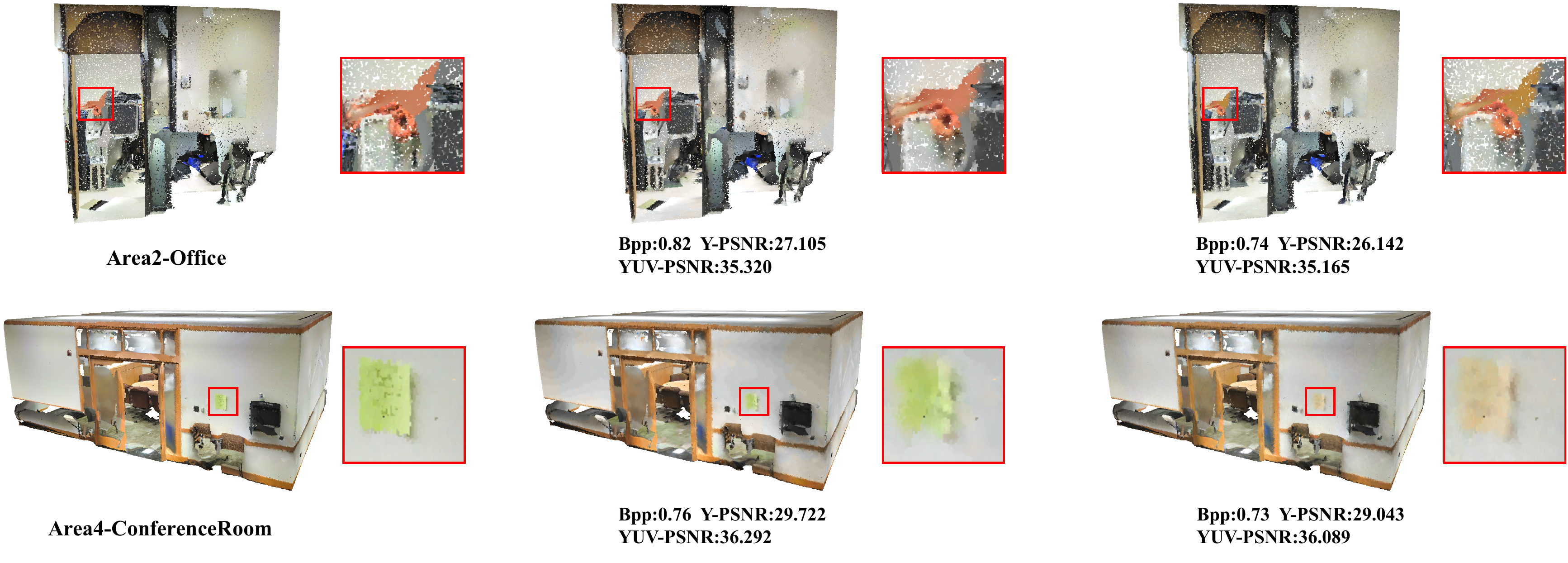}
\caption{Visual results of our method and Deep-PCAC on S3DIS. The left column is the original point cloud, the middle column is the point cloud reconstructed by our method, and the right column is the point cloud reconstructed by Deep-PCAC.}
\label{fig:s3dis_visual}  
\end{figure*}

\subsection{Subjective Evaluation}
We compare the test results of the proposed method with Deep-PCAC to illustrate the superiority of our approach in terms of subjective quality, as shown in Fig. \ref{fig:visual}, where our method demonstrates the higher visual quality at lower bit rates. For instance, the colorful area in legs and feet of the ``statue'' can be well reconstructed by our method, whereas Deep-PCAC degenerates the original color to gray artifacts. It can be also observed that our method preserves the green moss on the ``shiva'' point cloud decisively, yet Deep-PCAC clearly fails to reproduce the turquoise hue, even they are at a higher bit-rate budget. In the ``david9'' frame, Deep-PCAC generates blurred red stains in the character’s hairline, mouth, and nose, while our method maintains the same tone of color as the ground truth. And, in the ``phil9'' frame, in addition to the clearer contour of the human eyes, our method detects and restores the sparse green points in the original point cloud at the fingers, which Deep-PCAC fails to do so. 

Figure \ref{fig:s3dis_visual} shows the visualization results of the indoor scenes in S3DIS dataset, in which we can see that our method maintains basic color fidelity while Deep-PCAC suffers from shifted hues.

\subsection{Ablation Studies}

\begin{table*}[t]
\centering
\caption{ABLATION STUDY ON DIFFERENT RECEPTIVE FIELDS (COMPARE TO LAYER=2\&RATIO=4).}
\begin{tabular}{c|c|c|c|c|c|c|c|c}  
\toprule 
\multirow{2}{*}{\small{Point Cloud}}  & \multicolumn{2}{c|}{layers=1\&ratios=4}& \multicolumn{2}{c|}{layer=3\&ratios=4}& \multicolumn{2}{c|}{layer=2\&ratio=2}& \multicolumn{2}{c}{layer=2\&ratio=8}\\  \cline{2-9} 
& \footnotesize{BD-BR (\%)} & \footnotesize{BD-PR (dB)}& \footnotesize{BD-BR (\%)} & \footnotesize{BD-PR (dB)} & \footnotesize{BD-BR (\%)} & \footnotesize{BD-PR (dB)}& \footnotesize{BD-BR (\%)} &\footnotesize{BD-PR (dB)}\\ \hline
\small{Soldier}  & +296.38   &  -3.14  &  +92.51   &  -1.02 & +48.55    & -1.48     & \st{+1110.94}   & -1.74  \\ \hline
\small{Mask}     & \st{+2506.43}  & -1.89   &  \st{+1948.64} & -1.46  & +102.80   & -1.60     & \st{+2314.46}   & -1.66 \\ \hline
\small{Frog}     & \st{+3327.00}  & -3.52   &  +194.86  & -1.46  & +102.93   & -1.67     & \st{+3365.47}   & -2.32 \\ \hline
\small{House}    & +216.90   & -2.40   &  \st{+2578.91} & -1.39  & +52.73    & -1.39     & \st{+5343.49}   & -1.99  \\ \hline
\small{Shiva}    & +39.61    & -0.89   &  -73.92   & -0.81  & -27.28    & -0.89     & +3.29      & -1.93 \\ \hline
\small{Statue}   & +2.80     & -0.69   &  -99.98   & -3.39  & + 1.12    & -1.09     & +99.93     & -4.39 \\ \hline
\small{Andrew9}  & +96.26    & -2.79   & +82.26    & -1.22  &-40.33     & -1.57    &  -40.09    &  -1.43     \\ \hline
\small{Ricardo9} & +38.79    & -3.83   & -42.74    & -2.92  & +91.07    & -7.47    &  -44.26    &  -3.48   \\ \hline
\small{David9}   & +65.54    & -1.64   & -3.68     & -2.72  & +99.17    & -4.77     & -39.22     & -3.20  \\ \hline
\small{Phil9}    & +96.32    & -1.65   &  -39.60   &  -0.02 & +81.70    & -0.34     & -42.16     &+3.64    \\ \hline
\small{Area2}    & +279.63   & -2.64   & \st{+2532.19}  & -1.44  & +89.65    & -2.03     & \st{+3982.15}   &-2.15  \\ \hline
\small{Area4}    & +284.53   & -3.58   &  \st{+2149.23}  &  -0.99 & +69.75   & -1.44     & \st{+3523.25}   &-2.38 \\ \hline
\small{\textbf{Average}} & +118.06 &-2.38& -10.64  & -1.57  & +50.99   & -2.15    & -10.42   & -1.63 \\ \bottomrule
\end{tabular}
\label{tab:Different receptive fields}
\end{table*}

\begin{table}[t]
\centering
\caption{\textcolor{black}{ECA} COMPARED TO {NCA}.}
\begin{tabular}{c|c|c}  
\toprule 
\multirow{2}{*}{Point Cloud}  & \multicolumn{2}{c}{Naive Cross Attention}\\  \cline{2-3} 
& \footnotesize{BD-BR (\%)} & \footnotesize{BD-PSNR (dB)}  \\ \hline
Soldier  & -86.83    &   +3.24 \\ \hline
Mask     &  -88.32   & +1.68\\ \hline
Frog     &  - 96.785 & +1.71 \\ \hline
House    &  -41.547  &  +0.53 \\ \hline
Shiva    &  -169.08  & +1.11 \\ \hline
Statue   &  - 45.58  & +2.13 \\ \hline
Andrew9  & -36.54    & +1.75 \\ \hline
Ricardo9 & -98.65    & +1.53 \\ \hline
David9   & -65.84    & +1.07  \\ \hline
Phil9    & -67.95    & +1.81   \\ \hline
Area2    &  -89.54   & +1.75  \\ \hline
Area4    &  -91.64   & +1.89 \\ \hline
\textbf{Average}&  -74.16& +1.68 \\
\bottomrule
\end{tabular}
\label{tab:NCA}
\end{table}

\begin{table}[t]
\centering
\caption{ZERO-PADDING COMPARED TO FEATURE INTERPOLATION.}
\begin{tabular}{c|c|c}  
\toprule 
\multirow{2}{*}{Point Cloud}  & \multicolumn{2}{c}{feature interpolation}\\  \cline{2-3} 
& \footnotesize{BD-BR (\%)} & \footnotesize{BD-PSNR (dB)}  \\ \hline
Soldier  & +31.23   & +0.91  \\ \hline
Mask     &  -39.61  & +1.20 \\ \hline
Frog     & -24.51   & +1.09 \\ \hline
House    &  -61.24  & +0.92 \\ \hline
Shiva    &  -90.24  & +1.06 \\ \hline
Statue   &  -125.56 & +1.84 \\ \hline
Andrew9  & -36.10   & +1.04  \\ \hline
Ricardo9 & -29.54   & +0.89  \\ \hline
David9   & -36.27   &  +1.24 \\ \hline
Phil9    &  -26.58  & +0.92  \\ \hline
Area2    & -68.73   &  +0.75 \\ \hline
Area4    & -39.87   & +0.98  \\ \hline
\textbf{Average}& -50.79 & +1.07 \\
\bottomrule
\end{tabular}
\label{tab:int}
\end{table}

\begin{table}[t]
\centering
\caption{\textcolor{black}{COMPARISON OF EFFICIENCY}}
\begin{tabular}{c|c|c|c|c}  
\toprule 
\multirow{2}{*}{Point Cloud}  & \multicolumn{2}{c|}{\makecell{Deep-PCAC\\Model Size: 62.7MB} } & \multicolumn{2}{c}{\makecell{Ours\\Model Size: 39.3 MB}} \\ \cline{2-5} 
& \footnotesize{Encoding} & \footnotesize{Decoding} & \footnotesize{Encoding} & \footnotesize{Decoding}  \\ \hline
Soldier   &23.76    &14.24    &82.27   &17.45 \\ \hline
Mask      &8.19     &4.06     &19.26   &4.21  \\ \hline
Frog      &91.09    &60.06    &273.43  &57.67 \\ \hline
House     &100.75   &60.96    &367.42  &77.47 \\ \hline
Shiva     &22.21    &13.17    &75.23   & 15.86\\ \hline
Statue    &11.69    &6.88     &37.45   &7.85   \\ \hline
Andrew9   & 7.34    & 5.13    & 18.703 & 3.09   \\ \hline
Ricardo9  &11.36    &8.44     & 13.98  & 2.13   \\ \hline
David9    &8.61     &5.96     &19.33   & 5.26   \\ \hline
Phil9     &8.42     &5.26     & 21.19  & 4.01    \\ \hline
Area2     &1010.69  &961.55   &3015.2 & 600.04  \\ \hline
Area4     &1040.96 &547.54   &3045.19  &607.03 \\ \hline
\textbf{Average}& 195.51  & 141.18  & 582.38 & 116.68 \\
\bottomrule
\end{tabular}
\label{tab:time}
\end{table}

\subsubsection{Receptive Fields}
We employ a multi-scale approach for progressive downsampling and upsampling. This tactic allows the skeleton points within the point cloud to acquire an uniform receptive field size, enabling gradually integrating the color features from points within these receptive fields into compact feature representation.

Given that the receptive field size is impacted by the number of stacked sampling blocks and the sampling ratio, we conduct ablation study to analyze and determine the optimal receptive field size. As shown in Tab. \ref{tab:Different receptive fields}, we show ablation experiments for different number of stacked attention layers in \textcolor{black}{P-CA} module and sampling ratios of each sampling layer. Specifically, the following four cases is considered in this section: layers=1 \& ratio=4, layers=3 \& ratio=4, layers=2 \& ratio=2 and layers=2 \& ratio=8. We can observe that when layers=1 or layers=3, the performance of the model is clearly inferior to the layers number of 2. And, a sampling ratio of 4 can provide superior gains comparing with the ratio of 2 or 8. Experiments show that our model reports optimal compression efficiency when the number of sampling blocks is set to 2 and the sampling ratio is set to 4. 

\subsubsection{Naive-Cross Attention (NCA) v.s \textcolor{black}{ECA}}
When fusing geometry and attribute features, Naive Cross Attention (NCA) mechanism usually takes geometric coordinates as queries (Q) and attribute values as keys (K) and values (V).
In order to verify the superiority of our designed \textcolor{black}{External Cross Attention (ECA)} layer, we conduct related ablation study to compare with NCA.
According to Tab. \ref{tab:NCA}, when replacing \textcolor{black}{ECA} with NCA, the average BD-BR of the model is increased by 88.02\%, while the average BD-PSNR is reduced by 1.73dB, which shows that the proposed \textcolor{black}{ECA} is able to effectively fuse the attribute patterns with the geometry information.

\subsubsection{Feature Interpolation v.s Zero-Padding}
\textcolor{black}{On the decoding side, traditional methods often employ rule-based distance-weighted interpolation techniques to upsample attribute features.
In learning-based solutions~\cite{He_2023_CVPR,deepPCAC,sheng2022attribute}, the integration of linear layers with conventional methods has become prevalent.}
Instead, a direct and flexible zero-padding tactic is used in this paper, which requires the ablation study to verify the effectiveness of this tactic. 

As shown in Tab. \ref{tab:int}, replacing the zero-padding with feature interpolation leads to an average increase of 62.07\% in BD-BR, along with an average decrease of 1.17dB in BD-PSNR. It confirms zero-padding achieving lower bitrates while preserving higher reconstruction quality, as well as providing an interesting idea to up-sampling.

\subsection{\textcolor{black}{Comparison of Efficiency}}
To compare the efficiency of our proposed method and Deep-PCAC, we measure the encoding and decoding time on each point cloud of the test dataset. 
It can be seen from Tab. \ref{tab:time} that our method provides a faster decoding time than Deep-PCAC, which derives benefit from our zero-padding strategy that avoids redundant near-neighbor queries. 
However, we notice that the encoding time of our method is relatively long due to the attention mechanism. Although this result is influenced by the Python implementation, we believe that it does not hinder the potential of our approach.
Moreover, the model size of our method is 39.3 MB, i.e., 37\% smaller than Deep-PCAC, which not only implies lower memory usage and faster loading times but also indicates that our model can provide compression quality that is comparable to, or even superior than, Deep-PCAC while maintaining a smaller footprint. This finding highlights the compactness and efficacy of our method.
\textcolor{black}{Although the current encoding time is relatively long, this also points towards future research directions for improvement, such as reducing encoding time through algorithm optimization or parallel processing.}

\section{Conclusion}
Given the remarkable performance of attention mechanisms in point cloud analysis, this study introduces attention into the task of point cloud attribute compression. We propose a point cloud attribute compression network based on point-based model with an autoencoder architecture. Leveraging on a \textcolor{black}{ECA} module which effectively capture the correlation between geometry and color, our encoder can best exploit the local attribute pattern. At the decoding side, the attributes of the point cloud are progressively reconstructed based on the multi-scale representation and the zero-padding upsampling tactic. Experiments show that our proposed method, as a direct point-based approach, outperforms the state-of-the-art learning-based method Deep-PCAC. Although our method is still inferior to rules-based method, e.g., G-PCC, it brings fresh insights into the learning-based PCAC. 
 \textcolor{black}{In the future, we will focus on how to provide a time-efficient compression solution, which may include optimizing the attention mechanism to reduce computational complexity and facilitate  processing speed without compromising the reconstruction quality. This may involve exploring new network architectures, improving the efficiency of the attention modules, or incorporating advanced techniques such as quantization-aware training to minimize the computational overhead.}
\bibliographystyle{IEEEtran}
\bibliography{ref}

\end{document}